\documentclass[conference]{IEEEtran}
\IEEEoverridecommandlockouts
\usepackage{cite}
\usepackage{amsmath,amssymb,amsfonts}
\usepackage{graphicx}
\usepackage{textcomp}
\usepackage{xcolor}
\def\BibTeX{{\rm B\kern-.05em{\sc i\kern-.025em b}\kern-.08em
    T\kern-.1667em\lower.7ex\hbox{E}\kern-.125emX}}
\usepackage{epsfig}
\usepackage{color}
\usepackage{array}
\usepackage{comment}
\usepackage{multirow}
\usepackage{subcaption}
\usepackage{hhline}
\usepackage{algorithm}
\usepackage{wrapfig}
\usepackage{lipsum}
\usepackage[noend]{algpseudocode}
\usepackage{caption}
\newcommand{\argmin}{\operatornamewithlimits{argmin}}

\begin{document}

\title{Continual Learning with Support Boundary Experience Blending}

\author{
    \IEEEauthorblockN{
        Chih-Fan Hsu\textsuperscript{1},
	    Yung-Huei Chiou\textsuperscript{1},
	    Ming-Ching Chang\textsuperscript{2}, and 
	    Wei-Chao Chen\textsuperscript{1}
    }
    \IEEEauthorblockA{
        \textsuperscript{1}Inventec Corporation \\
        \textsuperscript{2}University at Albany, State University of New York \\
        \{hsuchihfan, chiouyh25\}@gmail.com, mchang2@albany.edu, weichao.chen@gmail.com
    }
}

\maketitle

\begin{abstract}

Continual learning (CL) seeks to mitigate catastrophic forgetting when models are trained with sequential tasks. A common approach, experience replay (ER), stores past exemplars but only sparsely approximates the data distribution, yielding fragile and oversimplified decision boundaries. We address this limitation by introducing {\bf Support Boundary Data (SBD)}, generated via differential-privacy-inspired noise into latent features to create boundary-adjacent representations that implicitly regularize decision boundaries. Building on this idea, we propose {\bf Experience Blending (EB)}, a framework that jointly trains on exemplars and SBD through a dual-model aggregation strategy. EB has two components: (1) latent-space noise injection to generate support boundary data, and (2) end-to-end training that jointly leverages exemplars and SBD. Unlike standard experience replay, SBD enriches the feature space near decision boundaries, leading to more stable and robust continual learning. Extensive experiments on CIFAR-10, CIFAR-100, Tiny ImageNet, and ImageNet-1k demonstrate consistent accuracy improvements of 10\%, 6\%, 14\%, and 9\%, respectively.


\end{abstract}


\section{Introduction}
\label{sec:intro}

Deep neural networks (DNNs) achieve remarkable performance across domains but typically rely on large, static datasets. In practice, however, data arrive sequentially, and models must adapt without retraining from scratch. Transfer learning~\cite{bao2023survey} offers a cost-effective way to fine-tune models on new data, but repeated fine-tuning leads to {\em catastrophic forgetting}---the rapid degradation of previously learned knowledge as new tasks are introduced. Continual Learning (CL)~\cite{Lange2022, Wang2023} addresses this challenge by enabling models to learn from a stream of tasks while retaining prior knowledge, thereby reducing performance loss on earlier tasks.


\begin{figure}[t]
\centerline{
\includegraphics[width=0.68\linewidth]{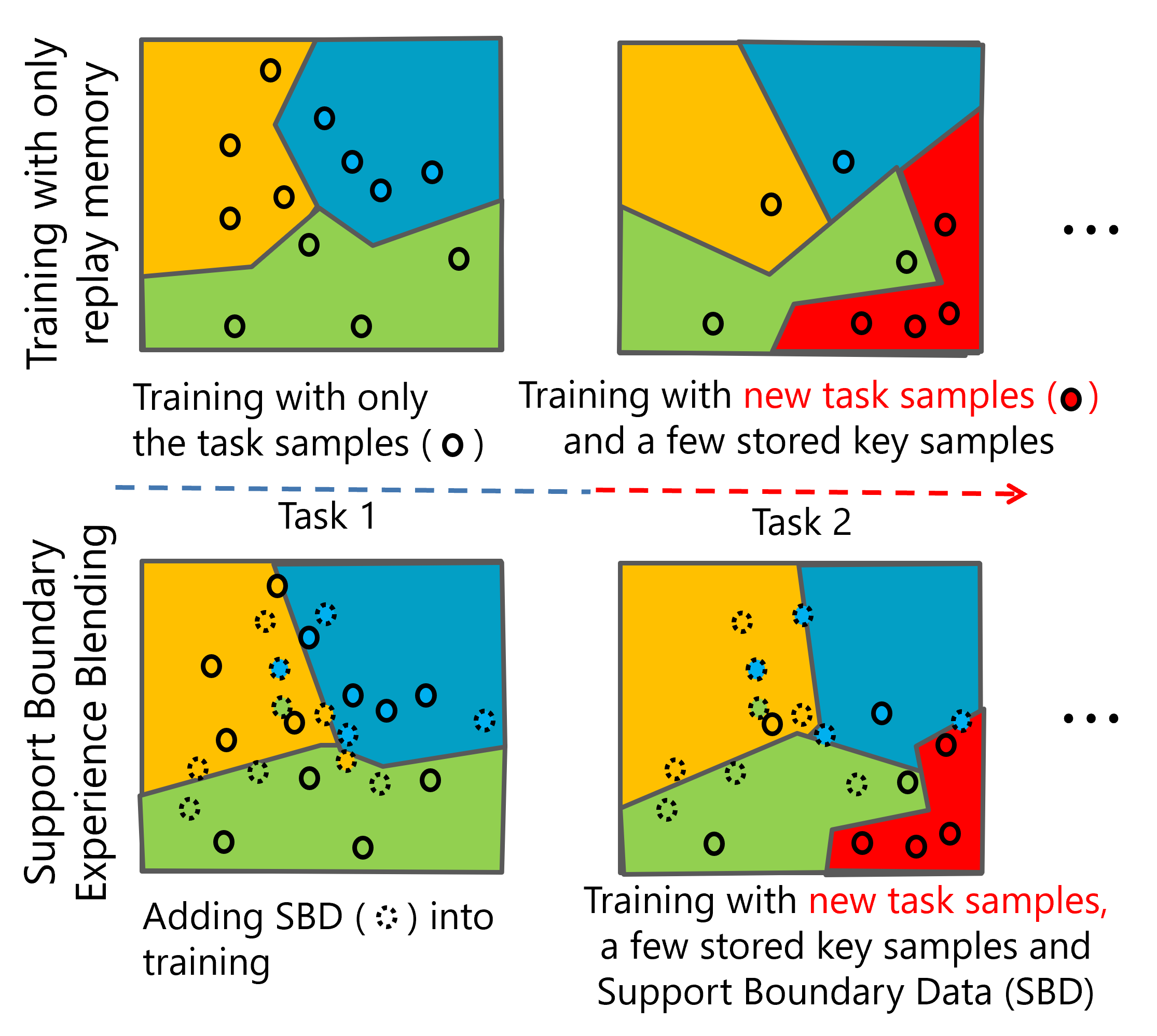}
}
\vspace{-6pt}
\caption{
Comparing our method with the ER method. We enhance the CL by memorizing the support boundary data (SBD) to improve decision boundary robustness. The red dots and region represent new task data and distribution.
}
\vspace{-10pt}
\label{fig:teaser}
\end{figure}

Experience Replay (ER) approaches have shown effectiveness in CL by retaining a limited set of key samples or class prototypes in replay memory. Specifically, ER methods select a small yet representative subset of exemplars that capture essential knowledge from previously learned tasks~\cite{Lange2022, Rebuffi2016iCaRLIC,9711397}.
Nevertheless, ER approaches still suffer from notable accuracy drops compared to standard learning, where the model is trained on all data simultaneously without the constraints imposed by CL. This decline in performance is primarily attributed to the domain shift between the replay memory samples and the full distribution of previously learned data. When the stored samples are sparsely distributed across the data domain, the resulting decision boundaries become overly simplistic, reducing the effectiveness of ER methods in mitigating forgetting. While increasing the size of the replay memory may offer some improvement, it does not fundamentally address the underlying problem, and the optimal memory size remains highly application-dependent.



{\em We hypothesize that introducing support data near the decision boundary during training acts as an implicit regularizer, stabilizing decision boundary update in CL and mitigating forgetting} (Figure~\ref{fig:teaser}). To achieve this idea, a key challenge is generating effective boundary data. Naive approaches, such as taking midpoints between class centroids, oversimplify the feature space and fail to produce representative boundary samples.
We introduce a method for generating boundary representations by injecting multivariate noise inspired by Differential Privacy (DP). While DP noise is traditionally used to obscure class identities and protect sensitive data, we repurpose it to create inherently ambiguous representations that naturally lie near decision boundaries. We refer to these representations as {\bf Support Boundary Data (SBD)}. 
Unlike Mixup or virtual adversarial training, which perturb in input or adversarial space, we inject DP-style Laplace noise into latent features to form boundary samples for CL.
Furthermore, we propose a novel training framework, {\bf Experience Blending (EB)}, which integrates knowledge from both stored key samples and support boundary data. By jointly leveraging key samples and SBD during CL training, we can form more stable decision boundaries and mitigate the effects of forgetting. Importantly, SBD is used only during training and does not incur additional computational cost during inference.





We conducted extensive experiments to evaluate our approach on four common public datasets: CIFAR-10, CIFAR-100~\cite{krizhevsky2009learning}, Tiny ImageNet~\cite{tiny-imagenet}, and ImageNet-1k~\cite{russakovsky2015imagenet}. The experimental results show that our method effectively mitigates forgetting across various CL settings, including the challenging Blurred Boundary Continual Learning (BBCL) setting, a combination of Class Incremental Learning (CIL) and highly non-IID Domain Incremental Learning (DIL). These findings support our hypothesis and highlight significant performance gains. Our approach consistently achieves significant performance gains on all datasets.

\section{Related Work}
\label{sec:related}

Continual Learning (CL) approaches aim to mitigate forgetting when a model is continuously trained on a stream of data~\cite{Lange2022}. Typically, this data stream $S$, is divided according to $T$ tasks, each containing a set of data batch $S^t_b =(X^t_b, Y^t_b)$, where $t\in \{1,\cdots T\}$, where $b$ is the batch index, $X$ represents the samples, and $Y$ represents the labels.
Catastrophic forgetting often leads to a significant decline in model performance due to domain shifts between tasks. An effective CL model should be robust to such distributional changes, maintaining consistent performance across all tasks. Consequently, the objective of CL is to minimize the loss function across all tasks and training batches, expressed as $\argmin_{\theta} \sum^{t}\sum^{b}L_{\theta}(X^t_b, Y^t_b)$, where $\theta$ is the model weights and $L$ is the loss.
Within the CL framework~\cite{Wang2023}, popular settings include Domain-Incremental Learning (DIL), Class-Incremental Learning (CIL), and Blurred Boundary Continual Learning (BBCL), each addressing catastrophic forgetting in different ways. 
Numerous methods mitigate forgetting by preserving trained knowledge. Among these methods, the experience replay (ER)-based and regularization-based approaches have shown outstanding performance.

{\bf ER-based approaches}: ER-MIR~\cite{Aljundi_2019_NueralIps} retrieves samples most affected by model updates. 
GDumb~\cite{prabhu2020greedy} uses a greedy approach to store samples in memory and retrains the model from scratch with these samples.
Rainbow Memory~\cite{bang_2021_cvpr} improves CL by managing memory based on per-sample classification uncertainty and data augmentation.
CLIB~\cite{koh2022online} retains essential data in a fixed-size memory and trains the model using this memory while adjusting the learning rate. In contrast, FMS~\cite{Lee_2023_ICCV} improves accuracy by using hierarchical class labels. However, the practicality of FMS may be constrained by the cost of obtaining precise annotations. FOSTER~\cite{wang2022foster} introduces feature boosting by freezing the model for previous tasks and adding a new module to combine features from both new and old samples.

{\bf Regularization-based approaches}: iCaRL~\cite{Rebuffi2016iCaRLIC} is a training strategy that utilizes knowledge distillation, prototype rehearsal, and nearest-mean-of-exemplars to improve sample representation and classification. BiC~\cite{Wu_2019_CVPR} employs bias correction and knowledge distillation to address distribution bias and manage incremental classes. EWC++~\cite{Chaudhry_2018_ECCV} applies KL-divergence-based regularization to preserve learned properties of previous tasks. DPCL~\cite{Lee_Oh_Chun_2024} perturbs the input and model weights with new memory management and adaptive learning rate to improve model generalizability.



Despite substantial progress in prior work, a significant performance gap persists compared to the upper bound (the performance of the standard training). Our method unifies experience replay and regularization by storing key samples alongside support boundary data (SBD). The latter serves as an implicit regularizer that encodes decision boundary information. Our novel Experience Blending (EB) framework further exploits SBD to support continual learning.



\section{Experience Blending Training Framework}
\label{sec:method}

\begin{figure}[t]
\centerline{
\includegraphics[width=0.80\linewidth]{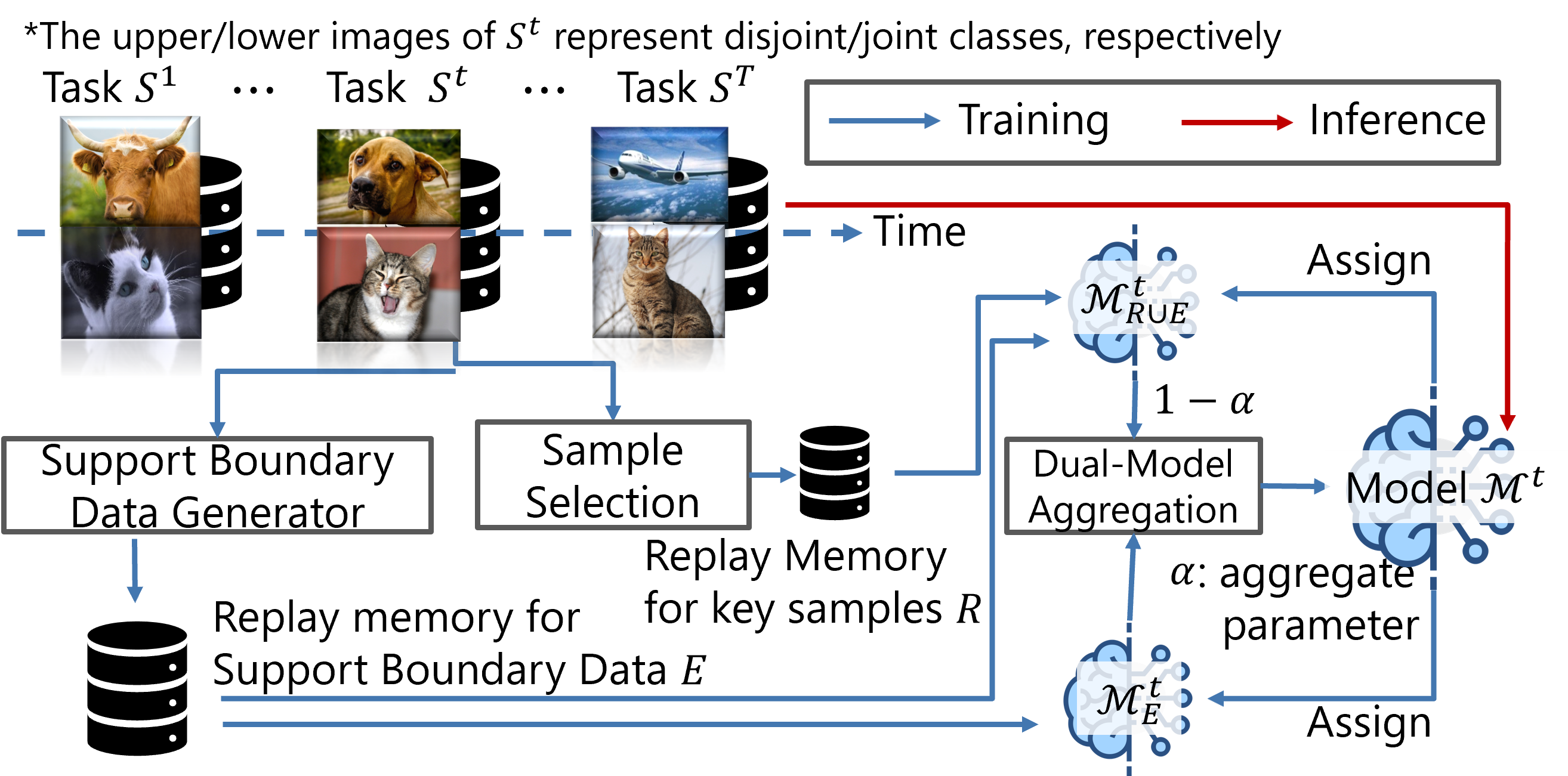}

}
\vspace{-6pt}
\caption{{\bf Experience blending}: Our method aggregates the model $\mathcal{M}_{R\cup E}$ trained on the key sample memory $R$ and the SBD memory $E$, and the model $\mathcal{M}_{E}$ trained only on $E$. The model receives only an image during inference.
}
\label{fig:Stages}
\vspace{-8pt}
\end{figure}

The experience blending training framework consists of two key components: (1) the generation of support boundary data (\S~\ref{sec:DEG}) and (2) an end-to-end training paradigm with a dual-model aggregation strategy (\S~\ref{sec:EB}).
The training and inference data flows are illustrated in Figure~\ref{fig:Stages}. 

\subsection{Support Boundary Data Generation}
\label{sec:DEG}


We generate data representations supporting class decision boundaries motivated by the concept of the latent noise-adding mechanism from differential privacy (DP)~\cite{dp2006}. It is important to note that the injected noise is not intended for privacy preservation, but rather serves as a form of regularization.

\noindent
{\bf Design Intuition.}
In DP, a Randomized Algorithm $RA(\cdot)$ satisfies  $(\varepsilon,0)$-DP if, for any pair of adjacent datasets $D$ and $D'$, and for every event $K$ in the output space:
$\mathbb{P}(RA(D)\in K)\leq e^{\varepsilon} \mathbb{P}(RA(D')\in K)$, where $\mathbb{P}(\cdot)$ denotes the probability distribution over outputs induced by the algorithm's randomness.
In practice, one common approach to achieving DP is by injecting noise (e.g., Laplace or Gaussian) based on the global sensitivity of a function. Specifically, this can be expressed as $RA(D) = \text{f}(D)+\frac{\Delta}{\varepsilon}Lap(0,1)$, where $\text{f}(\cdot)$ is the target function, $Lap(\cdot)$ denotes the Laplace distribution, and $\Delta$ is the global sensitivity of $\text{f}(\cdot)$. When applied to the latent representations, e.g., $\text{f}(D)=\mathcal{P}_R(D)$, where $\mathcal{P}_R$ is an arbitrary image encoder, the injected noise can perturb class-specific features, thereby reducing identifiability. Consequently, the perturbed representations may fall into more ambiguous regions of the feature space, where the model exhibits lower confidence. In some cases, the perturbation can even shift data points closer to the model’s decision boundaries. Figure~\ref{fig:latent} empirically illustrates this trend.

\begin{figure}[t]
\centerline{
\includegraphics[width=\linewidth]{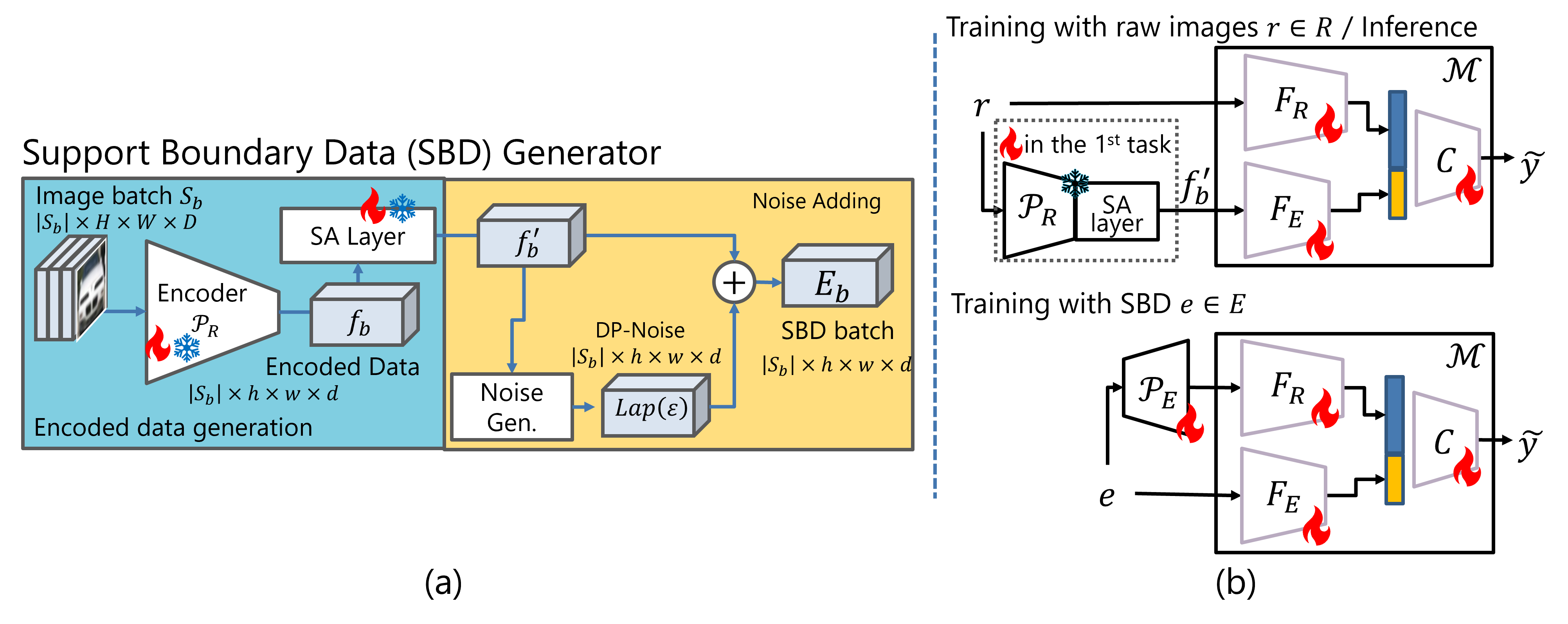}
}
\vspace{-6pt}
\caption{
(a) We generate SBD with an image encoder followed by an SA layer, where the DP-noise is added to the SA layer's output. The networks with ice or fire symbols indicate that the model is trainable or fixed. (b) The model $\mathcal{M}_{R\cup E}$ takes {\bf either} $r$ {\bf or} $e$ as input, while $\mathcal{M}_{E}$ only takes $e$. $F_R$ and $F_E$ are shared shallow subnetworks that integrate the inputs, and $C$ is the shared classifier.}
\label{fig:e_gen}
\vspace{-8pt}
\end{figure}

Building on this insight, we generate SBD via the process depicted in Figure~\ref{fig:e_gen}(a), which consists of two stages: {\em encoded data generation} and {\em noise adding}.



\noindent
{\bf Encoded Data Generation:}
We design the generation process with an image encoder $\mathcal{P}_R$ followed by a self-attention (SA) layer. 
An image batch $S^t_b$ from task $t$ is first passed through $\mathcal{P}_R$ to obtain encoded representations $f^t_b=\mathcal{P}_R(S^t_b)$. We omit $t$ here to reduce the notation in the equations. These representations are then processed by the SA layer, inspired by the Self-Attention GAN~\cite{pmlr-v97-zhang19d}, to emphasize class-specific features $f'_b$.
For more detail, in the SA layer, the $f_b \in \mathbb{R}^{|S_b|\times w\times h\times d}$ are projected into two feature spaces using matrices $W_\textrm{K}$ and $W_\textrm{Q}$, resulting in $\textrm{K}(f_b) = W_\textrm{K}f_b$ and $\textrm{Q}(f_b) = W_\textrm{Q}f_b$.
Then, $f'_b$ is obtained by:
$f'_{b,j} = \textrm{h}\left( \sum_{i=1}^{w\times h} \frac{\exp\left(\textrm{K}(f_{b,i})^{T}\textrm{Q}(f_{b,j})\right)\textrm{V}(f_{b,i})}{\sum_{i=1}^{w\times h}\exp\left(\textrm{K}(f_{b,i})^{T}\textrm{Q}(f_{b,j})\right)} \right)$,
where $i$ and $j$ indicate the location of the image patches; $\textrm{h}(\cdot)$ and $\textrm{V}(\cdot)$ are $1\times 1$ convolution layers.


\noindent
{\bf Noise Adding:}
Inspired by the Laplace noise-adding mechanism of DP, we perturb a batch of emphasized representations $f'_b$   with a Laplace distribution. Namely, the batch of SBD ($E_b$) is obtained by:
$E_b = f'_b + Lap
\left(
0, \frac{\max(f'_b) - \min(f'_b)}{\varepsilon|S^t_b|}
\right)$,
where $|S^t_b|$ is the batch size and $\varepsilon$ controls the magnitude of the multivariate noise. A small $\varepsilon$ value implies that the batch samples may become indistinguishable due to adding a large-variance noise. We have tested the $\varepsilon$ values with 0.001, 0.005, and 0.01. Overall, model performance remains stable for $\varepsilon \geq 0.005$, and we adopt this value as the default.


\subsection{Dual-Model Aggregation}
\label{sec:EB}
An important finding of our study is that incorporating SBD without further regularization is insufficient. Because SBD differs from the original data, the model tends to overfit to it, leading to memorization rather than generalization. As a result, SBD alone neither effectively prevents forgetting nor yields stable training. To counteract this overfitting and to better exploit both SBD and key samples, we adopt a Dual-Model Aggregation (DMA) strategy, which consistently improves both accuracy and training stability (rows 4 vs. 5 and 7 vs. 8 in Table~\ref{tab:abl_2}). DMA possesses two properties: (1) effective integration of knowledge from models trained on different data through weight aggregation, and (2) guaranteed convergence under arbitrary data distribution. These properties are particularly valuable in CL, where data from previous and new tasks are often non-i.i.d. and disjoint. We implement DMA using the federated aggregation algorithm of FedAvg~\cite{fedavg2017}, as its convergence has been theoretically established~\cite{fedavg_converge2020}, though the DMA can be readily extended to other aggregation methods.

\subsection{Experience Blending}
\label{sec:TF}
We refer to our method as Experience Blending because it integrates knowledge from stored exemplars and SBD. Although the DMA strategy draws inspiration from federated averaging, it is fundamentally different: the blending occurs between two internal models (replay+SBD vs. SBD-only), rather than across clients. This distinction allows EB to remain lightweight and specifically tailored for CL, as opposed to distributed training.
Let $S^t$ denote the task data of the task $t$, including data samples and corresponding labels. The process begins by generating and storing the SBD ($E^t$) for the samples of each task. The model $\mathcal{M}$ is then trained using the replay memory $R$ and the SBD memory $E$ for each task. We note that $S^t$ does {\it not} directly participate in model training. 
The replay memory $R$ is maintained with a sample selection module implemented by the sample-wise importance sampling described in~\cite{koh2022online}. According to this method, the least important sample is replaced when $R$ reaches its predefined maximum size. The importance of the sample is determined by the expected decrease in loss after updating the model.

\medskip
\noindent
{\bf Fine-tune the image encoder $\mathcal{P}_R$ in the first task.}
While a well-trained vision encoder can directly serve as the image encoder $\mathcal{P}_R$ for CL~\cite{IJCV2024}, the domain of the pre-trained model often differs from that of the downstream task. To address this domain gap, we adopt a {\em first-task fine-tuning} (FTF) strategy, inspired by First Session Adaptation~\cite{10378197}, to adapt $\mathcal{P}_R$ from the source domain to the target task domain. This fine-tuning is performed exclusively during the first task, as task data becomes inaccessible in subsequent tasks.
Specifically, we initialize $\mathcal{P}_R$ with a pre-trained model—such as one trained on ImageNet-1k or the vision encoder of CLIP~\cite{clip}—and fine-tune both $\mathcal{P}_R$ and the self-attention (SA) layer during the first task. Additionally, we update the stored support boundary data (SBD) at the end of each training epoch. For all subsequent tasks, both $\mathcal{P}_R$ and the SA layer are kept fixed to ensure stable and consistent feature encoding. Figure~\ref{fig:e_gen}(b) illustrates the data pipelines for two input types and highlights the trainable components during the first task. Overall, our method introduces no inference-time overhead and only a slight increase in training time, which depends on the size of the SBD memory (+0.13s per training epoch for 100 SBD samples).


\medskip
\noindent
{\bf Training Losses.}
Given a replay memory $R$ and a SBD memory $E$, training $\mathcal{M}$ with Experience Blending involves the use of two structurally very similar training processes. We first assign $\mathcal{M}$ to two models with identical structures, $\mathcal{M}_{R\cup E}$ and $\mathcal{M}_{E}$. $\mathcal{M}_{R\cup E}$ is updated with $R$ and $E$:
\begin{equation} \label{eq:mmix}
 L_{R\cup E} =  \mathcal{L}_{\{\mathcal{M}_{R\cup E},\mathcal{P}_R,\textrm{SA}(\cdot)\}}(r) + 
  \mathcal{L}_{\{\mathcal{M}_{R\cup E},\mathcal{P}_E\}}(e),  
\end{equation}
where $r\in R$, $e\in E$, $\mathcal{P}_E$ is a shallow transposed CNN that expands the dimension of the SBD to match that of the image, and $\mathcal{L}_\textbf{M}(\cdot)$ is the cross-entropy loss function for the set of models $\textbf{M}$. On the other hand, $\mathcal{M}_{E}$ is updated using only $E$:
\begin{equation} \label{eq:me}
 L_{E} = \mathcal{L}_{\{\mathcal{M}_{E},\mathcal{P}_E\}}(e), \text{where } e\in E.
\end{equation}
The final model $\mathcal{M}$ is a combination of $\mathcal{M}_{R\cup E}$ and $\mathcal{M}_E$, controlled by an aggregate parameter $\alpha$. Namely, 
$\mathcal{M} = (1-\alpha) \mathcal{M}_{R\cup E} + \alpha \mathcal{M}_{E}$, where $\alpha=0.5$.

\subsection{How SBD Samples Support Decision Boundary Formation}

\begin{figure}[t]
\begin{center}
\includegraphics[width=\linewidth]{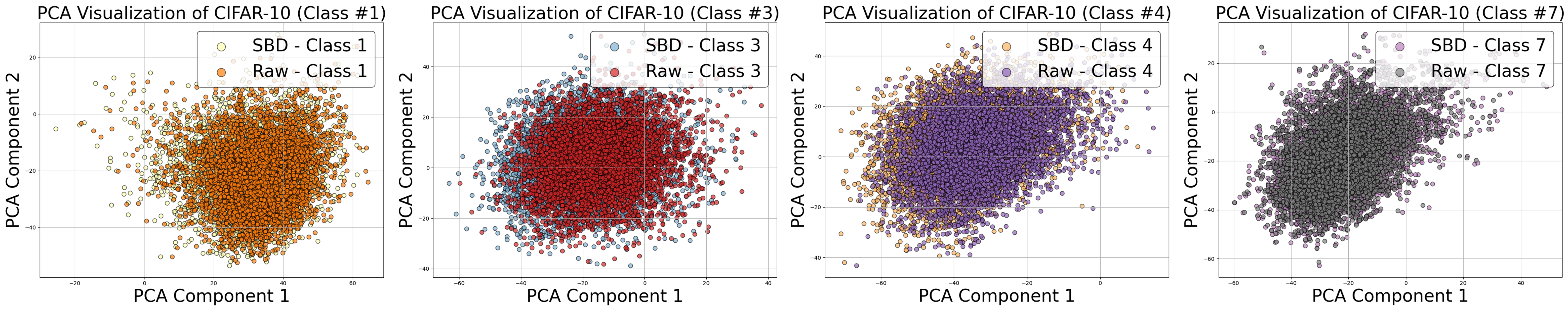}
\end{center}
\vspace{-12pt}
\caption{{\bf Latent visualization}. The dispersion of SBD beyond the raw data distribution suggests that they are effectively positioned near decision boundaries.}
\label{fig:latent}
\end{figure}

We examine how Support Boundary Data (SBD) supports the formation of decision boundaries by analyzing both its geometric behavior and its quantitative effect on sample margins.
To obtain an intuitive view, we apply Principal Component Analysis (PCA) to project both SBD samples and the latent representations of raw CIFAR-10 samples into a two-dimensional space. As shown in Figure~\ref{fig:latent}, SBD samples consistently extend beyond the core distribution of the raw data, forming outward shifts relative to the class distributions. Although PCA is a low-dimensional projection, the systematic nature of this shift indicates that the noise-injection mechanism does not merely create random perturbations; rather, it effectively generates support samples positioned closer to the decision boundary. This provides visual, empirical evidence that SBD enriches the boundary-adjacent regions of the data distribution.

To complement this qualitative observation, we quantitatively evaluate the effect of SBD by measuring the margin~\cite{jiang2019} of raw samples under different module configurations. The margin is defined as
$\text{Margin}(x) = f_{y}(x) - \max_{k \neq y} f_k(x)$,
where $f_y(x)$ denotes the logit corresponding to the ground-truth label $y$, and $f_k(x)$ represents the logit of class $k$. Larger positive margin values indicate greater separation from the decision boundary, and negative values indicate misclassification. As reported in Table~\ref{tab:margin}, incorporating SBD and EB into training notably increases the average margin while significantly reducing its variance across three random seeds. This reflects a more stable decision boundary.

Taken together, our qualitative and quantitative analyses show that SBD acts as training-only support samples populating the boundary regions---some lying between different domain regions---thereby enlarging the decision margin during CL training. This guides the model toward more robust, better-defined decision boundaries, which in turn improves downstream classification accuracy.

\begin{table}[t]
\centering
\caption{Statistics of CIFAR-100 raw samples across three random seeds.}
\label{tab:margin}
\resizebox{\columnwidth}{!}{
\begin{tabular}{lrrr}
\hline
 & \multicolumn{1}{c}{FTF} & \multicolumn{1}{c}{EB (w/o DMA)} & \multicolumn{1}{c}{EB} \\ \hline\hline
Average Accuracy ($A_{avg}$) ($\uparrow$) & 30.25$_{\pm0.36}$ & 53.98$_{\pm0.35}$ & 59.06$_{\pm0.17}$  \\ \hline
Average of Margin ($\uparrow$) & -0.40$_{\pm0.17}$ & 2.02$_{\pm0.24}$ & 5.21$_{\pm0.04}$  \\ \hline
Variance of Margin ($\downarrow$) & 23.03$_{\pm1.54}$ & 18.59$_{\pm0.37}$ & 9.50$_{\pm0.09}$ \\ \hline
\end{tabular}
}
\vspace{-6pt}
\end{table}






\section{Experimental Results}
\label{sec:exp}

We evaluate the performance of our method on four datasets at different scales: CIFAR-10, CIFAR-100, Tiny ImageNet, and ImageNet-1k. All experiments are conducted on a Linux-based container equipped with a 4-core AMD EPYC 7543 CPU, 128 GB RAM, and an Nvidia A40 GPU. The maximum size $|R|$ is set to 500, 2k, 4k, 20k for CIFAR-10, CIFAR-100, Tiny ImageNet, and ImageNet-1k, respectively. All tested methods share the same $|R|$, unless explicitly stated otherwise. For ImageNet-1k, both $|E|$ and the $|R|$ for our method are set to 10k for fair memory usage. In our experiments, we use a ResNet-18 model pre-trained on ImageNet-1k, provided by PyTorch, as the default encoder $\mathcal{P}_R$ initialization. We also initialize  $\mathcal{P}_R$ with a CLIP~\cite{clip} in some experiments to demonstrate the extensibility of our method. A batch size of 128 is used across all datasets, with a default configuration of 5 tasks, unless stated otherwise. Each task is trained for 10 epochs using a learning rate of $0.01$.


\medskip
\noindent
{\bf The training/validation/test sets:}
We use the i-Blurry-$n$-$m$ setup from~\cite{koh2022online} to partition the training and test sets. In this setup, $n$\% of classes are {\em disjoint}, meaning these class samples are uniquely assigned to specific tasks based on their labels. The remaining (100-$n$)\% are {\em blurry}. Samples from these classes are distributed across all tasks with a specified blurry level $m$. This blurry level $m$ controls the degree of data distribution imbalance within a task; a smaller $m$ indicates a severe non-i.i.d. task data. Unless otherwise specified, we use default values of $n=50$ and $m=10$.

\begin{figure}[t]
\begin{center}
\includegraphics[width=\linewidth]{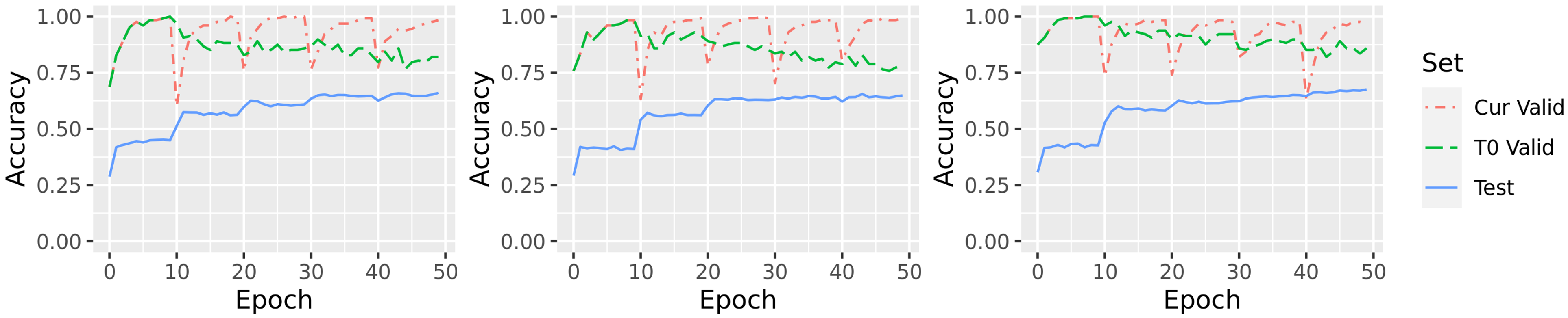}
\end{center}
\vspace{-12pt}
\caption{Validation and test accuracy plots.}
\label{fig:lc}
\end{figure}
\begin{table}[t]
\footnotesize
\caption{{\bf Comparison of average accuracy} $A_{avg}$ (\%) across the datasets using the i-Blurry-50-10 setting. The gray rows show the prior works. Numbers following $\pm$ represent the standard deviation across three experiments with different seeds. Bold and italics indicate the 1$^{st}$ and 2$^{nd}$ highest $A_{avg}$.
}
\label{tab:Aavg}
\vspace{-10pt}
\begin{center}
\resizebox{\columnwidth}{!}{
\begin{tabular}{lrrrrr}
\hline
\multicolumn{1}{c}{Dataset} & \multicolumn{1}{c}{CIFAR-10} & \multicolumn{1}{c}{CIFAR-100}  & \multicolumn{1}{c}{Tiny ImageNet} & \multicolumn{1}{c}{ImageNet-1k} \\ 
\multicolumn{1}{l}{Method} & \multicolumn{3}{c}{($t$=$5$,$\mathcal{P_R}$=IN)} & \multicolumn{1}{c}{($t$=$10$,$\mathcal{P_R}$=CLIP)} \\ \hline
{\color{gray} Standard}   &  {\color{gray}96.03} & {\color{gray}79.89}                & {\color{gray}53.05}     & {\color{gray}69.26} \\ \hline
{\color{gray}RM~\cite{bang_2021_cvpr}}            &  {\color{gray}61.52$_{\pm3.69}$}      & {\color{gray}33.27$_{\pm1.59}$}       & {\color{gray}17.04$_{\pm0.77}$}      &  {\color{gray} 28.30 ($|R|$=$20k$)}  \\ \hline
{\color{gray}GDumb~\cite{prabhu2020greedy}}         &  {\color{gray}55.27$_{\pm2.69}$}      & {\color{gray}34.03$_{\pm0.89}$}       & {\color{gray}18.69$_{\pm0.45}$}      &  {\color{gray}14.21 ($|R|$=$20k$)}   \\ \hline
{\color{gray}EWC++~\cite{Chaudhry_2018_ECCV}}         &  {\color{gray}60.33$_{\pm2.73}$}      & {\color{gray}38.78$_{\pm2.32}$}       & {\color{gray}24.39$_{\pm1.18}$}     &  {\color{gray} 26.21 ($|R|$=$20k$)}   \\ \hline
{\color{gray}ER-MIR~\cite{Aljundi_2019_NueralIps}}        &  {\color{gray}61.93$_{\pm3.35}$}      & {\color{gray}38.28$_{\pm1.15}$}       & {\color{gray}24.54$_{\pm1.26}$}     &  {\color{gray} 20.68 ($|R|$=$20k$)}   \\ \hline
FTF        &  62.32$_{\pm0.42}$      & 30.25$_{\pm0.36}$       & 24.58$_{\pm0.22}$  & 22.50    \\ \hline
{\color{gray}BiC~\cite{Wu_2019_CVPR}}           &  {\color{gray}61.49$_{\pm0.68}$}      & {\color{gray}37.61$_{\pm3.00}$}       & {\color{gray}24.90$_{\pm1.07}$}      &  {\color{gray} 28.38 ($|R|$=$20k$)} \\ \hline
{\color{gray}CLIB~\cite{koh2022online}}          &  {\color{gray}73.90$_{\pm0.22}$} & {\color{gray}49.22$_{\pm0.79}$}       & {\color{gray}25.05$_{\pm0.52}$}      &  {\color{gray} 28.88 ($|R|$=$20k$)}\\ \hline
{\color{gray}iCaRL~\cite{Rebuffi2016iCaRLIC}}         &  {\color{gray}68.77$_{\pm2.88}$}      &  {\color{gray}33.55$_{\pm0.58}$}      & {\color{gray}25.41$_{\pm0.55}$}       &  {\color{gray} --} \\ \hline
{\color{gray}FOSTER~\cite{wang2022foster}}        &  {\color{gray}73.40$_{\pm1.20}$}      & {\color{gray}52.80$_{\pm0.15}$}  & {\color{gray}33.93$_{\pm0.47}$}  &  {\color{gray} 33.62$^\dagger$}\\ \hline
{\color{gray}DPCL$^*$~\cite{Lee_Oh_Chun_2024}}         &  {\color{gray}--}      &  {\color{gray}50.22$_{\pm0.39}$}      & {\color{gray}--}     &  {\color{gray} --}   \\ \hline
EB (w/o DMA)                       &  \it{83.14$_{\pm0.59}$} & \it{53.98$_{\pm0.35}$}  & \it{40.21$_{\pm0.99}$} & \it{35.01} ($|R|$=$|E|$=$10k$)\\ 
EB (w/ DMA)                         &  \bf{84.35$_{\pm1.06}$} & \bf{59.06$_{\pm0.17}$}  & \bf{47.68$_{\pm0.22}$} & \bf{42.77} ($|R|$=$|E|$=$10k$)\\ \hline
\end{tabular}
}
\end{center}
\vspace{-5pt}
\footnotesize{*The DPCL code has not been released. The accuracy is reported in Table 2 of~\cite{Lee_Oh_Chun_2024}. $^\dagger$The training takes more than 125 hours.}\\
\vspace{-20pt}
\end{table}

\subsection{Overall Evaluation}

We evaluate our method by comparing it with nine common CL methods, including iCaRL~\cite{Rebuffi2016iCaRLIC}, EWC++~\cite{Chaudhry_2018_ECCV}, BiC~\cite{Wu_2019_CVPR}, ER-MIR~\cite{Aljundi_2019_NueralIps}, GDumb~\cite{prabhu2020greedy}, RM~\cite{bang_2021_cvpr},  FOSTER~\cite{wang2022foster}, CLIB~\cite{koh2022online}, and DPCL~\cite{Lee_Oh_Chun_2024}, in terms of average accuracy $A_{avg}$ (\%), computed from three experimental runs with different random seeds.
Here, $A_{\text{avg}}$ measures the average accuracy of the previously encountered classes in the test set at the end of each task, and is defined as $A_{avg} = \frac{1}{T}\sum_{t=1}^{T}a_t$, where $a_t$ is the accuracy for task $t$ and $T$ is the number of tasks. We emphasize that this comparison is not intended to claim the superiority of our method; rather, it aims to show that the use of SBD can contribute to improvements in $A_{\text{avg}}$.

\begin{table}[t]
\footnotesize
\caption{{\bf Comparison of $A_{avg}$ in three task configurations} on CIFAR-100 under the i-Blurry-50-10 setting. Our method outperforms the compared methods.
}
\label{tab:tasks}
\vspace{-10pt}
\begin{center}
\begin{tabular}{lrrr}
\hline
       & \multicolumn{1}{c}{5 Tasks} & \multicolumn{1}{c}{10 Tasks} & \multicolumn{1}{c}{25 Tasks} \\ \hline
{\color{gray}iCaRL~\cite{Rebuffi2016iCaRLIC}}  & {\color{gray}33.55$_{\pm0.58}$}& {\color{gray}24.21$_{\pm0.50}$}               & {\color{gray}14.57$_{\pm0.90}$}               \\ \hline
{\color{gray}EWC++~\cite{Chaudhry_2018_ECCV}}  & {\color{gray}38.78$_{\pm2.32}$}& {\color{gray}34.85$_{\pm1.71}$}               & {\color{gray}29.46$_{\pm1.65}$}               \\ \hline
{\color{gray}ER-MIR~\cite{Aljundi_2019_NueralIps}} & {\color{gray}38.28$_{\pm1.15}$}& {\color{gray}35.29$_{\pm1.56}$}               & {\color{gray}30.00$_{\pm1.17}$}               \\ \hline
{\color{gray}GDumb~\cite{prabhu2020greedy}}  & {\color{gray}34.03$_{\pm0.89}$}& {\color{gray}32.35$_{\pm1.73}$}               & {\color{gray}31.87$_{\pm1.08}$}               \\ \hline
{\color{gray}FOSTER~\cite{wang2022foster}} & {\color{gray}52.80$_{\pm0.15}$}& {\color{gray}42.59$_{\pm1.25}$}               & {\color{gray}33.17$_{\pm1.22}$}               \\ \hline
{\color{gray}BiC~\cite{Wu_2019_CVPR}}    & {\color{gray}37.61$_{\pm3.00}$}& {\color{gray}37.40$_{\pm1.04}$}               & {\color{gray}35.55$_{\pm1.38}$}               \\ \hline
{\color{gray}RM~\cite{bang_2021_cvpr}}    & {\color{gray}33.27$_{\pm1.59}$}& {\color{gray}34.93$_{\pm3.94}$}               & {\color{gray}36.85$_{\pm1.17}$}               \\ \hline
{\color{gray}CLIB~\cite{koh2022online}}   & {\color{gray}49.22$_{\pm0.79}$}& {\color{gray}46.15$_{\pm1.07}$}          & {\color{gray}\it{43.64$_{\pm1.08}$}}          \\ \hline
EB (IN)   & \it{59.06$_{\pm0.17}$}& \it{50.27$_{\pm2.99}$}          & 38.46$_{\pm1.30}$          \\ \hline
EB (CLIP)   & \bf{59.99$_{\pm1.32}$}& \bf{56.69$_{\pm5.65}$}          & \bf{57.44$_{\pm0.67}$}          \\ \hline
EB (trained on $S$) & {61.70$_{\pm0.15}$} & {59.63$_{\pm0.54}$} & {59.51$_{\pm0.15}$} \\ \hline
\end{tabular}
\end{center}
\vspace{-12pt}
\end{table}

Table~\ref{tab:Aavg} shows the results of initializing $\mathcal{P}_R$ with ImageNet for CIFAR-10/100 and Tiny ImageNet, and with CLIP for ImageNet-1k. The `Standard' row shows the $A_{avg}$ when the model is trained without CL setting (upper-bound). Overall, our method outperforms all baselines, with gains of 10.45\% on CIFAR-10, 6.26\% on CIFAR-100, 13.75\% on Tiny ImageNet, and 9.15\% on ImageNet-1k. Due to the long training time, certain ImageNet-1k experiments could not be completed; we have therefore left these entries blank in our results. The `FTF' row in the table reports the results of applying first-task fine-tuning, which serves to align the domain of the pre-trained model with that of the downstream task. However, despite fine-tuning the well-trained image encoder, the improvement in $A_{avg}$ remains limited. The experiments at the bottom of the table underscore the effectiveness of incorporating SBD and the dual-model aggregation (DMA) strategy. In general, directly integrating SBD into model training already outperforms all baseline methods. The DMA strategy further improves the $A_{avg}$ for all evaluated datasets. These results demonstrate that the DMA strategy enables effective utilization of SBD.
We further evaluate the level of forgetting by common evaluation metrics, average forgetting, BWT, and FWT~\cite{Lp2017}. Based on test accuracy, our method achieves -3.5$_\pm$0.5\% of average forgetting, 14.0$_\pm$2.0\% of BWT, and 40.5$_\pm$0.5\% of FWT, demonstrating its effectiveness in mitigating forgetting and generalizing to future tasks.

\noindent
{\bf The Number of Tasks.}
To evaluate the generalizability of our method across different pre-trained encoders and different numbers of tasks, we compare three initialization strategies for the encoder $\mathcal{P}_R$: using the model pre-trained on ImageNet (IN), using the CLIP model (ViT-B/32), and using the model trained on entire dataset of the downstream task $S$ as the performance upper-bound. Table~\ref{tab:tasks} presents the $A_{avg}$ results for continual learning with 5, 10, and 25 tasks.
In the ImageNet initialization experiments, we observe that performance drops by approximately 8.79\% when increasing from 5 to 10 tasks, and by 20.6\% when increasing to 25 tasks.  This decline is primarily due to the limited and biased data coverage in the first task, which constrains the encoder’s ability to generalize across subsequent tasks. A more powerful encoder can mitigate the effects of limited early-task data, as evidenced by the experiments using the CLIP model, where the degradation in accuracy is significantly reduced.


\subsection{Controlling Memory Budget}


\noindent
{\bf Controlling the Replay Memory Size $|R|$.}
We examine the effect of replay memory size ($|R|$) on both average accuracy ($A_{avg}$) and final task accuracy ($A_{fin}$). With $|R|$ set to 500, 1,000, and 2,000, the corresponding $A_{avg}$ values are 57.57${_\pm0.60}$, 58.24${_\pm0.20}$, and 59.06${_\pm0.17}$, while $A_{fin}$ improves from 64.84${_\pm1.15}$ to 65.11${_\pm0.47}$ and 66.51${_\pm0.97}$, respectively.
We observe a positive correlation between $|R|$ and accuracy. Notably, our method outperforms all compared methods by 4.7\%, even with only 500 stored samples. In comparison, FOSTER, which ranks second with an accuracy of 52.80\%, requires storing 2,000 samples. The relatively small differences in $A_{avg}$ and $A_{fin}$ across varying memory sizes highlight the effectiveness of our method in leveraging knowledge from the SBD.
\begin{figure}[t]
 \clearpage
        \begin{center}
        \includegraphics[width=0.5\linewidth]{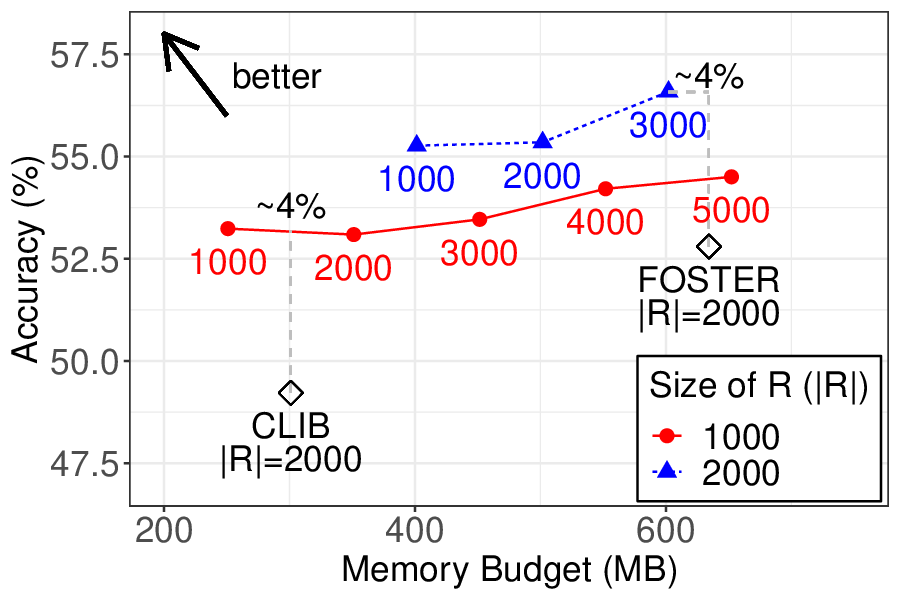}
        \end{center}
        \vspace{-12pt}
        \caption{This figure shows how varying replay memory size $|R|$ (represented by the color curves) and the SBD memory size $|E|$ (indicated by the numbers below the points) impacts accuracy under different memory budgets.}
        \label{fig:budget}
        \vspace{-6pt}
\end{figure}


\noindent
{\bf Impact on the SBD Memory Size $|E|$.}
The size of each SBD point is primarily defined by its dimensionality and numerical precision. Ideally, the framework would store as much SBD as possible without limitations. In this section, we evaluate our approach under a restricted memory budget by constraining $|E|$ with a sample selection mechanism similar to that used for $R$. We set $|R|$ from 1,000 to 2,000 and vary $|E|$ from 1,000 to 4,000 on CIFAR-100 to assess the impact of available memory sizes. For baseline comparisons of $A_{avg}$, CLIB and FOSTER with $|R|$=$2,000$, where CLIB achieves $A_{avg}$ 49.22\% with 301.1 MB of memory usage (storing raw images), while FOSTER achieves 52.80\% using 634.3 MB, storing raw images and four frozen encoders. As shown in Figure~\ref{fig:budget}, our method achieves approximately  4\% improvement over CLIB and FOSTER under the same memory budgets, demonstrating more efficient memory utilization. The trade-off between memory allocation to $R$ and $E$ and its effect on model accuracy is a promising direction for further investigation.

\subsection{Robustness on Different CL Settings}

Table~\ref{tab:data_settings} evaluates how different data distributions across tasks affect CIFAR-100. A higher $m$ value indicates less skewed data distributions for blurry classes, while a larger $n$ value means more disjoint class distributions across tasks.
We also compare the final accuracy $A_{fin}$ (\%). This metric reflects both overall accuracy and forgetting levels. The first three rows show that increasing $m$ improves both $A_{avg}$ and $A_{fin}$, as expected since less skewed data distributions generally enhance model performance. However, the improvement from less-skewed data shows diminishing returns.

The bottom five rows of Table~\ref{tab:data_settings} analyze the effect of the disjoint ratio $n$. The results indicate that a blurry setting with a low blurry level is more challenging than a fully disjoint setting. This finding is consistent with the CIFAR-10 results reported in the CLIB paper. The lower $A_{avg}$ in blurry settings is due to the slow adaptation to blurry classes, as evidenced by the difference between $A_{avg}$ and $A_{fin}$.
In the CIL setting, $A_{avg}$ reaches 77.64\%, reflecting the ease of learning due to i.i.d. data and the limited number of 20 classes per task. However, $A_{avg}$ declines over time during training, with $A_{fin}$ dropping to 64.12\%, highlighting this trend.

Even compared to the top-performing method, FOSTER, which achieves $52.80\%$ $A_{avg}$ with $n$=$50$, our method shows superior performance with $53.85\%$ $A_{avg}$ when $n$=$25$ in the BBCL setting. Additionally, $A_{fin}$ remains stable across all settings. Overall, our method demonstrates better generalizability across varying data distributions.

\begin{table}[t]
\footnotesize
\caption{{\bf Comparison of different data distributions:} $m$ represents the blurry level and $n$ denotes the disjoint ratio. Smaller values of $n$ or $m$ indicate a more challenging setting.
}
\label{tab:data_settings}
\vspace{-12pt}
\footnotesize
\begin{center}

\begin{tabular}{cccrr}
\hline
\multicolumn{1}{c}{Setting} &\multicolumn{1}{c}{$n$} & \multicolumn{1}{c}{$m$} & \multicolumn{1}{c}{$A_{avg}$} & \multicolumn{1}{c}{$A_{fin}$} \\ \hline\hline
BBCL& 50  & 10         &   59.06$_{\pm0.17}$ &  66.51$_{\pm0.97}$ \\ \hline
BBCL& 50  & 30         &   66.57$_{\pm0.23}$ &  67.83$_{\pm0.51}$ \\ \hline
BBCL& 50  & 50         &   68.84$_{\pm0.49}$ &  69.09$_{\pm0.48}$ \\ \hline \hline
DIL &  0  & 10         &   50.38$_{\pm0.51}$ &  64.93$_{\pm1.09}$ \\ \hline
BBCL& 25  & 10         &   53.85$_{\pm0.57}$ &  66.15$_{\pm0.35}$ \\ \hline
BBCL& 50  & 10         &   59.06$_{\pm0.17}$ &  66.51$_{\pm0.97}$ \\ \hline
BBCL& 75  & 10         &   64.61$_{\pm0.67}$ &  65.62$_{\pm0.50}$ \\ \hline
CIL &100  & --         &   \multicolumn{1}{l}{77.64} &   \multicolumn{1}{l}{64.12} \\ \hline
\end{tabular}
\end{center}
\vspace{-10pt}
\end{table}

\subsection{Ablation Study}
\label{sec:AS}

We conducted an ablation study to assess the impact of our design choices on model performance, measured by $A_{avg}$, using the i-Blurry-50-10 setting on CIFAR-100 and Tiny ImageNet. In the experiments, the encoder $\mathcal{P}_R$ was initialized with a model pre-trained on ImageNet. Table~\ref{tab:abl_2} summarizes the results across various design combinations. Specifically, {\bf FTF} indicates that $\mathcal{P}_R$ is fine-tuned using the first task’s data; {\bf R} denotes the use of key sample replay memory during training; {\bf SBD} refers to the incorporation of support boundary data; and {\bf DMA} represents the use of the dual-model aggregation strategy. In all settings, the model $\mathcal{M}$ is still trained, even when individual components are disabled.

\begin{table}[t]
\caption{Impact on Various Design Combinations}
\vspace{-10pt}
\label{tab:abl_2}
\begin{center}
\footnotesize
\begin{tabular}{ccccrr}
\hline
\multicolumn{4}{c}{Design} & \multicolumn{2}{c}{$A_{avg}$}                                                          \\ \hline
FTF   & R & SBD &DMA  & \multicolumn{1}{c}{CIFAR-100}            & \multicolumn{1}{c}{Tiny IN}        \\ \hline\hline
  -   &     -     &  -  &  - & 24.81$_{\pm0.16}$                        & 20.62$_{\pm0.12}$                        \\ 
\checkmark     &    -      &  - &  - & 30.25$_{\pm0.36}$                        & 24.58$_{\pm0.22}$                        \\ 
\checkmark     & \checkmark         &   - &  - & 42.20$_{\pm0.91}$                        & 24.92$_{\pm0.30}$                        \\ 
\checkmark     & -       &  \checkmark &  -  & 47.95$_{\pm3.56}$                        & 42.32$_{\pm4.05}$                        \\ 
\checkmark     & -       &  \checkmark &  \checkmark  & 52.22$_{\pm1.60}$                        & 46.97$_{\pm1.83}$                        \\ 
\multicolumn{4}{c}{{\color{gray}FOSTER~\cite{wang2022foster}}} & {\color{gray}52.80$_{\pm0.15}$} & {\color{gray}33.93$_{\pm0.47}$} \\
\checkmark     & \checkmark       &  \checkmark &  -  & 53.98$_{\pm0.35}$                        & 40.21$_{\pm0.99}$                        \\ 
\checkmark     & \checkmark         & \checkmark &  \checkmark  & \bf{59.06$_{\pm0.17}$}        & \bf{47.68$_{\pm0.22}$}        \\ \hline

\end{tabular}
\end{center}
\vspace{-12pt}
\end{table}


The results reveal the following insights:
(1) The improvement in $A_{avg}$ is limited when using the pre-trained $\mathcal{P}_R$ without FTF. This is because the features extracted by the pre-trained encoder are not well aligned with the downstream task, even when the pre-training domain is similar (e.g., ImageNet vs. Tiny ImageNet);
(2) Incorporating key sample replay memory into CL training improves $A_{avg}$, though the extent of improvement varies by dataset (+12\% on CIFAR-100 vs. only +0.4\% on Tiny ImageNet). In contrast, the inclusion of SBD consistently benefits CL performance on both datasets; and
(3) Using only FTF, key sample replay memory, and SBD already outperforms the strong baseline---FOSTER. Adding the DMA strategy further enhances the model’s resistance to forgetting during CL training. Overall, Experience Blending effectively integrates SBD to CL training, leading to substantial performance gains in both datasets.
\section{Conclusion}

We introduced Support Boundary Data (SBD) to mitigate catastrophic forgetting in continual learning (CL). Our Experience Blending (EB) framework integrates stored exemplars with SBD, stabilizing decision boundaries and improving knowledge retention. Experiments on multiple benchmarks, including challenging Blurred Boundary CL settings, show substantial and consistent performance gains, underscoring the practicality and robustness of the approach.

{\bf Limitations:} The EB framework currently assumes well-defined task boundaries for first-task fine-tuning, which limit applicability in task-free scenarios. Although SBD generation is more efficient than adversarial perturbations, it introduces modest training overhead. Moreover, the optimal balance between exemplar memory and SBD memory remains unexplored, highlighting the need for adaptive allocation strategies.

\bibliographystyle{IEEEtran}
\bibliography{main}

\end{document}